\documentclass[11pt]{article}

\usepackage[preprint]{acl}

\usepackage{times}
\usepackage{latexsym}

\usepackage[T1]{fontenc}

\usepackage[utf8]{inputenc}

\usepackage{microtype}

\usepackage{inconsolata}

\usepackage{graphicx}
\usepackage{subcaption}
\usepackage{amsmath}
\usepackage{amssymb}
\usepackage{booktabs}
\usepackage{booktabs,siunitx,makecell,multirow}

\usepackage{algorithm}
\usepackage{algpseudocode}
\title{Unlocking the Black Box of Latent Reasoning: An Interpretability-Guided Approach to Intervention}

\author{
\textbf{Shuochen Chang\textsuperscript{1,*}}, \textbf{Tong Bai\textsuperscript{2,*}}, \textbf{Xiaofeng Zhang\textsuperscript{1}}, \textbf{Qianli Ma\textsuperscript{1}},\\
\textbf{Qingyang Liu\textsuperscript{1}}, \textbf{Zhaohe Liao\textsuperscript{1}}, \textbf{Yibo Miao\textsuperscript{1}}, \textbf{Li Niu\textsuperscript{1,$\dagger$}} \\[0.3em]
\textsuperscript{1}Shanghai Jiao Tong University \hspace{1em}
\textsuperscript{2}Fudan University \\
\texttt{\{csc1332741686,framebreak,mqlqianli\}@sjtu.edu.cn} \\
\texttt{tbai22@m.fudan.edu.cn} \\
\texttt{\{narumimaria,zhaoheliao,miaoyibo,ustcnewly\}@sjtu.edu.cn} \\[0.2em]
\small{\textsuperscript{*}Equal contribution \hspace{1em} \textsuperscript{$\dagger$}Corresponding author}
}

\begin{document}
\maketitle

\begin{abstract}
Latent reasoning enables Large Language Models (LLMs) to perform multi-step inference within continuous hidden states, offering efficiency gains over explicit Chain-of-Thought (CoT). However, the opacity of these continuous thought vectors hinders their reliability and controllability. This paper bridges the gap between mechanistic interpretability and actionable control. We first present a systematic analysis using structural, causal, and geometric probes, revealing that latent vectors encode compressed, faithful representations of reasoning steps, with early vectors acting as critical causal hubs. Building on this, we operationalize these interpretability insights into a suite of training-free, decode-time interventions that refine the latent reasoning process by imposing the identified geometric and semantic priors. Extensive experiments across multiple model scales and diverse task domains  demonstrate that our approaches consistently improve reasoning accuracy. Our interpretability-guided interventions consistently unlock latent capabilities and improve reasoning accuracy without any parameter updates.
\end{abstract}

\section{Introduction}

Large Language Models (LLMs) have evolved from simple pattern completion to performing sophisticated multi-step reasoning~\citep{deepseekr1,gpt4_tech}. Much of this progress stems from techniques like Chain-of-Thought (CoT) prompting~\citep{cot}, where models generate explicit, token-based rationales to structure their inference process. While effective, this reliance on verbalized reasoning introduces significant drawbacks. Explicit CoTs can be excessively long~\citep{cou1,d3tom,liu2026break}, incurring substantial latency and computational costs. Furthermore, their textual nature provides a low-bandwidth medium for complex computation~\citep{coconut}, often containing redundant information and increasing susceptibility to compounding errors~\citep{cot_faithfulness,dramaqa}.

These limitations motivate a shift toward latent reasoning~\citep{survey1,survey2,survey3}, where multi-step inference unfolds within the model's continuous hidden states without generating intermediate text. This approach enables more efficient computation by bypassing the need for explicit token generation. Existing methods explore various paradigms, such as optimizing hidden state trajectories~\citep{colar,soft,seek_dark,quiet_star} or reusing network layers~\citep{loop1,loop2,dep1,sys15} to simulate deeper computation. Among these, the continuous-thought paradigm~\citep{coconut,codi,pretrain,simcot,marcos}, which employs special latent vectors to propagate information across reasoning steps. It has shown particular promise due to its simplicity and compatibility with standard decoding pipelines.

Despite its potential, the continuous-thought paradigm faces two critical gaps in understanding. First, its interpretability remains largely unverified: it is unclear whether these latent vectors genuinely encode semantic steps of a reasoning process or are merely artifacts of the training objective~\citep{internalization_deng}. Second, existing studies are typically limited to smaller-scale models, leaving the scalability and behavior of this paradigm at larger scales under-explored. This paper aims to bridge these gaps: Not by proposing a new training method, but by establishing rigorous principles and diagnostic tools for interpreting and intervening in the latent reasoning process.

To address the first gap whether latent thought vectors genuinely encode semantic reasoning steps, we develop a suite of three complementary interpretability probes~\citep{simcot,latent_sft,rep_eng,linear_probe}, designed to examine their structure, content, and causal role in the reasoning process. First, through structural alignment analyses~\citep{cka,cka_revisit,othello_gpt}, we find that the geometry of latent thought vectors closely mirrors that of explicit CoT rationales, suggesting a shared representational scaffold.
Second, we show that a simple linear map can reconstruct textual reasoning steps from these latent vectors with high fidelity, confirming they serve as compressed yet faithful encodings of intermediate reasoning. Finally, causal intervention experiments demonstrate that targeted edits to these vectors especially in early reasoning steps to predictably alter the model’s final answer, establishing their functional role in the inference pipeline~\citep{rome,activation_steering,reft,inference_time_alignment,contrastive_decode}.

Building on these insights, we design a series of training-free, decode-time interventions. By manipulating cached latent states by projecting them along directions identified by our probes, we can both validate our interpretability hypotheses and measurably improve reasoning performance without any parameter updates. These consistent gains from simple, norm-preserving edits highlight the untapped potential residing within the model’s existing latent pathways.

Our contributions are threefold:
\begin{enumerate}
    \item \textbf{Unified Interpretability Framework:} We provide the first systematic analysis of the continuous-thought paradigm, establishing structural, pointwise, and causal evidence that latent vectors are not opaque artifacts but steerable, semantic reasoning blueprints.
    \item \textbf{From Theory to Practice:} We translate our interpretability findings into a novel suite of training-free, decode-time interventions. By enforcing semantic consistency and geometric regularity, we demonstrate that model performance can be improved solely by manipulating inference dynamics.
    \item \textbf{Scalable and Robust Validation:} We extend the evaluation of latent reasoning to large-scale models (up to 8B) and diverse domains. Our results on GSM8K, OOD benchmarks, and StrategyQA confirm that the identified causal mechanisms are fundamental and generalizable properties of latent reasoning.
\end{enumerate}

\begin{figure*}[t]
  \centering
  \includegraphics[width=\textwidth]{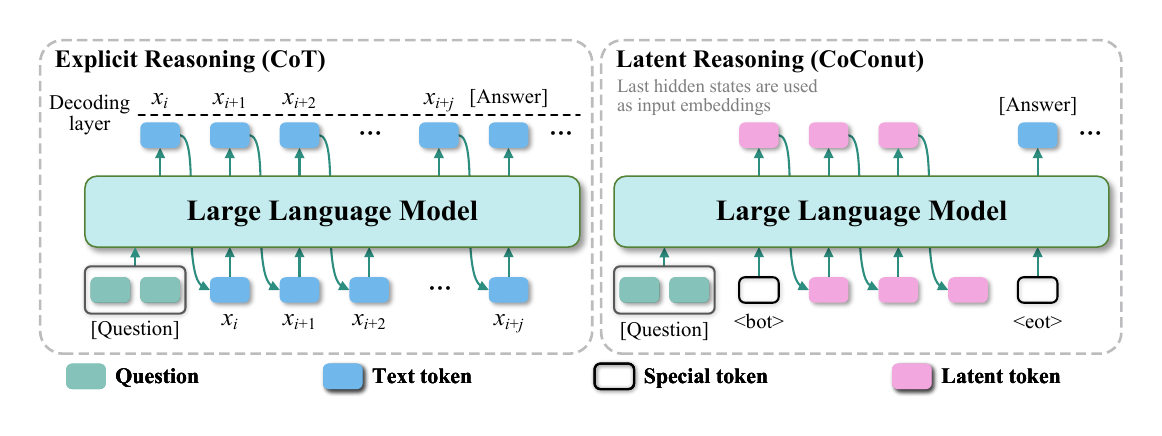}
  \caption{\textbf{Explicit Reasoning versus Latent Reasoning}. Left: Explicit Reasoning externalizes intermediate reasoning as a token sequence (\(x_i, x_{i+1}, \ldots, x_{i+j}\)).Right: Latent Reasoning reserves a latent span and at each reserved slot, feeds the preceding final-layer hidden state back as the next input, yielding a continuous thought that conditions subsequent decoding.}
  \label{fig:cot-vs-latent-wide}
\end{figure*}

\section{Related Works}

\subsection{Paradigms for Latent Reasoning}
\label{sec:rw-latent}
Latent reasoning seeks to realize multi-step inference inside hidden states, avoiding explicit intermediate tokens while retaining standard decoding for the final answer. Prior work advances mechanisms as follows. Latent optimization refines hidden trajectories in place~\citep{loop1,loop2,lta_thinker,pause_token} during decoding or distills long textual rationales into compact continuous representations. Signal-guided control steers trajectories with auxiliary signals such as confidence estimates, verifier outputs, or decode-time selection criteria, and may bias states toward informative embedding subspaces~\citep{quiet_star,camp_fire,thought_retriever,bu3}. Layer-recurrent~\citep{dep1,sys15,latent_evolve,bu1} execution emulates deeper computation without lengthening the sequence by reusing layers or activations, for example through architectural loops or by feeding a layer’s output back as the next input. Our study focuses on the continuous-thought latent-span variant~\citep{simcot,soft,pretrain,coconut,marcos}, in which reserved latent vectors are written into the prefix to carry information across steps, as popularized by CoConut-style models. Those most demonstrations target small models and their scalability and interpretability at larger scales remain underexplored.

\subsection{Interpretability of Latent Reasoning}
\label{sec:rw-interpret}
Interpretability efforts examine where and how hidden computation acquires stepwise structure and causal influence. Mechanistic analyses~\citep{simcot,latent_sft,zhao2026,marcos,lta_thinker} track the emergence of staged computation within transformer internals, relating attention flow and layer specialization to progressive subgoal formation. Behavioral studies~\citep{seek_dark,colar,step_internalization,bu2,bu3} infer latent processing from macroscopic signatures such as abrupt memorization-to-generalization transitions and the internalization of intermediate steps that manifest as step skipping at inference. Representational analyses~\citep{codi,explain1,latent_evolve,step_skip,sae_features} link hidden trajectories to explicit reasoning via centered kernel alignment, linear decoding, and lightweight mappers, and corroborate these relations through targeted causal edits that ablate, steer, or transplant states~\citep{reft,iti}. Our method contributes the interpretability by integrating structural alignment, then operationalizing the resulting priors as training-free, decode-time interventions. The protocol both supports interpretability and yields consistent gains in reasoning accuracy, connecting explanation with capability improvement.

\section{Interpretation of Latent Reasoning}
\label{sec:latent-interpretation}

To investigate whether latent reasoning genuinely performs structured, multi-step inference, we conduct a three-part analysis. We first probe the semantic content of the latent vectors by measuring their correspondence with explicit Chain-of-Thought (CoT) representations (Section~\ref{sec:probe}). Next, we perform a series of causal interventions to test whether manipulating these vectors predictably controls the model's output (Section~\ref{sec:causal-latents}). Finally, we explore the architectural and geometric properties that enable this emergent computational structure (Section~\ref{sec:arch-geom}).

\subsection{Preliminary}
\label{sec:prelim}

Our work focuses on a mode of reasoning where inference occurs within the model's continuous hidden states, rather than through verbalized CoT. A standard causal language model with parameters $\theta$ defines the probability of a sequence $w_{1:T}$ as:
\begin{equation}
p_\theta(w_{1:T}) = \prod_{t=1}^{T} p_\theta(w_t \mid w_{<t}).
\end{equation}
While explicit CoT involves minimizing cross-entropy over a sequence of textual reasoning steps, latent reasoning reserves a special span of $K$ positions within the prompt that do not correspond to vocabulary tokens. The computation within this span is realized through a sequence of hidden vectors we term continuous thoughts.

Formally, given an input embedding matrix $\mathbf{E} \in \mathbb{R}^{T \times d}$, a latent span of length $K$ is introduced. The model deterministically populates this span by sequentially generating $K$ continuous thought vectors, $\mathbf{z}_{1:K}$. At each latent position $\ell_k$, the vector is derived from the previous position's final hidden state and then used as the input embedding for the current position:
\begin{equation}
\mathbf{z}_k \triangleq \mathbf{h}^{(L)}_{\ell_k-1}, \quad \text{and} \quad \tilde{\mathbf{E}}_{\ell_k} \leftarrow \mathbf{z}_k,
\end{equation}
where $\mathbf{h}^{(L)}_{t}$ is the last-layer hidden state at position $t$ and $\tilde{\mathbf{E}}$ is the resulting "filled" embedding matrix. Conditioned on $\tilde{\mathbf{E}}$, the model proceeds with standard autoregressive decoding to generate the final answer $y$:
\begin{equation}
p_\theta(y \mid x) \equiv p_\theta\bigl(y \mid \tilde{\mathbf{E}}(x)\bigr).
\end{equation}
Crucially, the latent vectors $\mathbf{z}_k$ are not directly supervised. Their values are learned implicitly through the standard next-token prediction loss, which is applied only to the visible target tokens $\mathcal{V}_{\mathrm{vis}}$ (i.e., the answer):
\begin{equation}
\label{eq:loss}
\mathcal{L}(\theta) = - \sum_{t \in \mathcal{V}_{\mathrm{vis}}} \log p_\theta\big( w_{t} \mid w_{<t};\, \tilde{\mathbf{E}}(x) \big).
\end{equation}

\subsection{Probing the Semantic Content of Latent Vectors}
\label{sec:probe}

Our first hypothesis is that the sequence of continuous thoughts, $\mathbf{z}_{1:K}$, serves as a compressed, continuous analogue to a textual CoT. To test this, we assess the correspondence between the latent representations and the hidden states generated by an equivalent model executing an explicit CoT.

\begin{figure*}[t]
    \centering
    \includegraphics[width=0.9\textwidth]{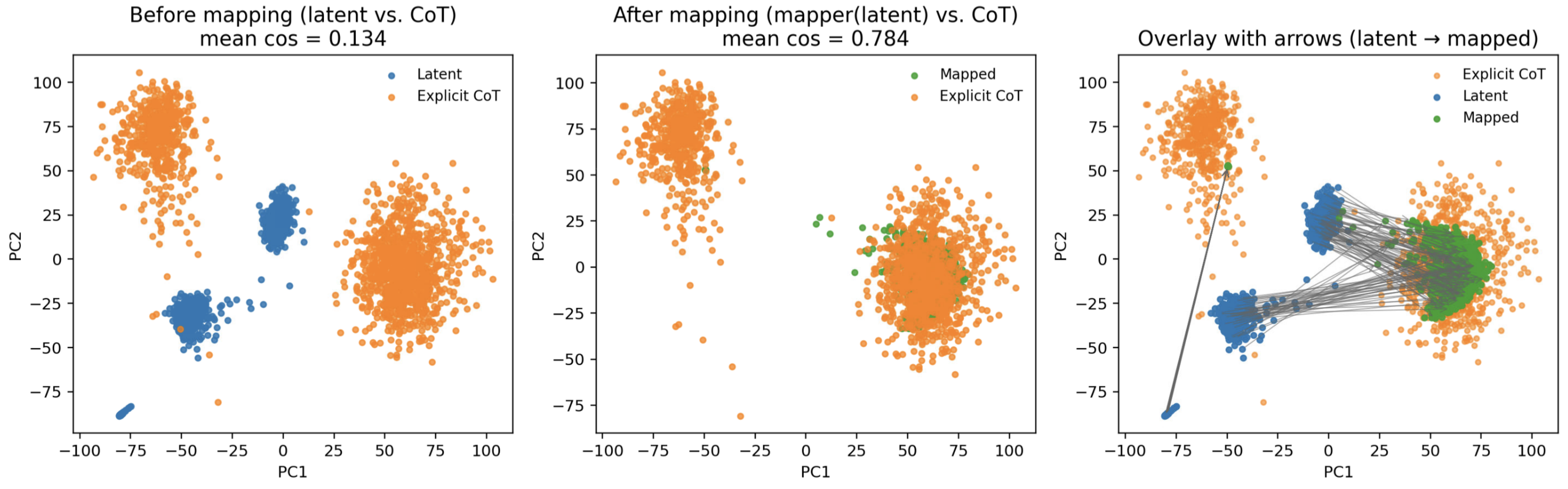}
    \caption{\textbf{Mapper-based alignment of latent thoughts to explicit CoT}. (a) Before mapping, latent and CoT points are separated in PCA space. (b) A linear mapper \(f_{\phi}\) aligns latents with CoT clusters. (c) Overlay with arrows (latent\(\rightarrow\)mapped) visualizes the systematic displacement toward CoT.}
    \label{fig:procrustes}
\end{figure*}

\paragraph{Establishing Correspondence.}
Let $\mathcal{T}$ be the set of textual step boundaries in a teacher-forced CoT. We extract the final-layer hidden state at the end of each step $t \in \mathcal{T}$ to serve as its canonical representation: $\mathbf{c}_{t}(x) \triangleq \mathbf{h}^{(L)}_{\mathrm{end}(t)}(x)$. To create a comparable latent representation, we locally average a small, contiguous block of continuous thoughts that are heuristically aligned with step $t$, forming a step-level latent surrogate:
\begin{equation}
\label{eq:avg-latent}
\bar{\mathbf{z}}_{t}(x) \triangleq \frac{1}{|\mathcal{K}(t)|} \sum_{k \in \mathcal{K}(t)} \mathbf{z}_k(x).
\end{equation}
This aggregation allows us to test for step-wise correspondence without making strong assumptions about a one-to-one mapping between latent vectors and reasoning steps.

\paragraph{Multi-faceted Probing Analysis.}
We employ three complementary methods to probe the relationship between the latent surrogates $\{\bar{\mathbf{z}}_t\}$ and their explicit counterparts $\{\mathbf{c}_t\}$.
First, we measure their \textbf{structural alignment} using linear Centered Kernel Alignment (CKA), which quantifies the similarity of the geometric arrangement of data points across different instances. CKA is defined as:
\begin{equation}
\label{eq:cka}
\mathrm{CKA}(\tilde{\mathbf{Z}}_{t}, \tilde{\mathbf{C}}_{t}) = \frac{\|\tilde{\mathbf{Z}}_{t}^{\top}\tilde{\mathbf{C}}_{t}\|_{F}^{2}}{\|\tilde{\mathbf{Z}}_{t}^{\top}\tilde{\mathbf{Z}}_{t}\|_{F} \cdot \|\tilde{\mathbf{C}}_{t}^{\top}\tilde{\mathbf{C}}_{t}\|_{F}},
\end{equation}
where $\tilde{\mathbf{Z}}_{t}$ and $\tilde{\mathbf{C}}_{t}$ are matrices whose rows are the centered latent surrogates and CoT step representations, respectively, for a set of problems.

Second, we test for pointwise recoverability by training a simple linear map $f_{\phi}$ to predict a CoT representation $\mathbf{c}_t(x)$ from its corresponding latent surrogate $\bar{\mathbf{z}}_t(x)$. The model is trained to maximize cosine similarity, a stringent test of linear decodability:
\begin{equation}
\label{eq:cosloss}
\mathcal{L}_{\mathrm{cos}}(\phi) = \mathbb{E}_{t,x} \left[1 - \cos\bigl(f_{\phi}(\bar{\mathbf{z}}_{t}(x)), \mathbf{c}_{t}(x)\bigr)\right].
\end{equation}
Finally, we conduct a non-parametric lexical probe to see if a latent surrogate contains information about the tokens in the corresponding reasoning step. We feed $\bar{\mathbf{z}}_t(x)$ through the model's language modeling head and measure the probability mass assigned to the top tokens of the actual CoT step:
\begin{equation}
\label{eq:lex-probe}
p_{\mathrm{vocab}}(\cdot \mid \mathbf{z}) = \mathrm{softmax}\bigl(\mathbf{W}\times\mathrm{LN}(\mathbf{z})+\mathbf{b}\bigr).
\end{equation}

\paragraph{Results.} 
Table~\ref{tab:semantic-probes} confirms a high degree of semantic fidelity. The strong CKA scores indicate a significant geometric alignment between latent and explicit reasoning spaces, while the high cosine similarity verifies that CoT steps are linearly recoverable. Additionally, the lexical probe demonstrates robust token-level decodability. As visualized in Figure~\ref{fig:procrustes}, these results establish that latent vectors function not as opaque artifacts, but as faithful, compressed encodings of the reasoning process.

\begin{table}[t]
\centering
\small
\setlength{\tabcolsep}{4pt}
\begin{tabular}{lcc}
\toprule
\textbf{Metric} & \textbf{Value} & \textbf{\small{vs. Baseline}} \\
\midrule
\multicolumn{3}{l}{\textit{\textbf{I. Structural Alignment} (Geometry)}} \\
Linear CKA Score & \textbf{0.72} & -- \\
\addlinespace[4pt]
\multicolumn{3}{l}{\textit{\textbf{II. Pointwise Recoverability} (Content)}} \\
Cosine Similarity & \textbf{0.75} & \small{+0.61 (Identity)} \\
\addlinespace[4pt]
\multicolumn{3}{l}{\textit{\textbf{III. Lexical Probe} (Vocabulary)}} \\
Top-1 Token Accuracy & \textbf{0.38} & \small{> Random} \\
\bottomrule
\end{tabular}
\caption{\textbf{Semantic Correspondence Analysis.} Latent vectors exhibit high structural similarity (CKA) and linear recoverability relative to explicit CoT steps, confirming they are compressed semantic states.}
\label{tab:semantic-probes}
\end{table}

\subsection{Analyzing the Causal Role of Latent Thought Sequences}
\label{sec:causal-latents}
Having established a correlational link between latent vectors and reasoning steps, we now investigate their causal role. If continuous thoughts truly guide the reasoning process, then targeted interventions on them should predictably alter the model's final answer. We measure the effect of an intervention $\mathcal{I}$ on a latent sequence $\mathbf{Z}$ by the change in log-probability of the target answer $\mathbf{y}^{\star}$:
\begin{equation}
\label{eq:delta-logp}
\Delta\log p_{\text{tgt}} \triangleq \log p(\mathbf{y}^{\star}|\mathcal{I}(\mathbf{Z})) - \log p(\mathbf{y}^{\star}|\mathbf{Z}).
\end{equation}

\paragraph{Intervention Strategies.}
We probe causal influence through three complementary intervention families applied to cached latent vectors. To assess positional salience, we truncate the latent sequence after the $k$-th vector, denoted $\mathrm{drop}@k$, thereby removing downstream stages of latent computation and measuring the resulting change in prediction. To test fine-grained controllability, we introduce small, norm-preserving edits to an early latent state: a spherical interpolation that reorients the vector toward an answer-supporting boundary (Eq.~\ref{eq:s1}), and a single normalized gradient step that nudges the vector in the direction that increases the likelihood of the correct answer (Eq.~\ref{eq:s2}). To evaluate transferability, we transplant the entire latent sequence from one instance into another and examine whether the model’s answer shifts toward the donor’s target, treating the sequence as a putative internal reasoning plan. Together, these interventions jointly characterize where information is concentrated, whether it can be steered with minimal edits, and to what extent latent trajectories encode reusable guidance for decoding.

\begin{figure}[t]
  \centering
  \includegraphics[width=0.9\columnwidth]{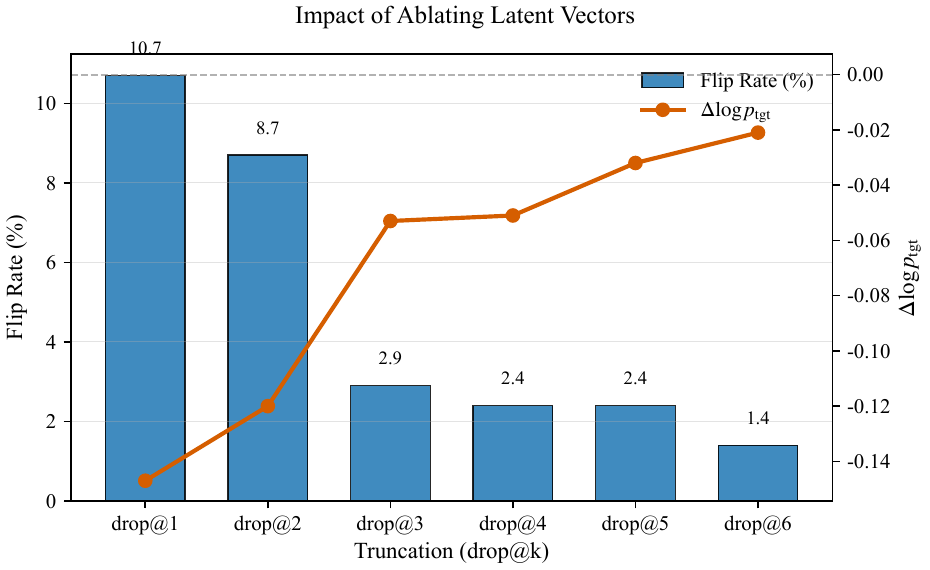}
  \caption{Impact of ablating latent vectors. The answer flip rate decreases as more of the initial latent vectors are preserved (from drop@0 to drop@K), confirming that early latent thoughts are causally critical for the final outcome.}
  \label{fig:ablation}
\end{figure}

\begin{table}[t]
\centering
\small
\setlength{\tabcolsep}{2.5pt}
\begin{tabular}{@{}l c c c@{}}
\toprule
\textbf{Intervention Type} & \textbf{Flip \%} & \textbf{$\Delta\log p$} & \textbf{Dir. \%} \\
\midrule
\multicolumn{4}{l}{\textit{A. Ablation (Positional Importance)}} \\
\hspace{3mm} Drop@0 (Remove all) & 10.7 & -0.15 & -- \\
\hspace{3mm} Drop@1 (Keep first) & 8.7 & -0.12 & -- \\
\addlinespace[4pt]
\multicolumn{4}{l}{\textit{B. Targeted Edits (Steerability)}} \\
\hspace{3mm} Slerp (to Answer) & 21.5 & \textbf{+0.85} & -- \\
\hspace{3mm} Gradient Update & 24.1 & \textbf{+1.02} & -- \\
\addlinespace[4pt]
\multicolumn{4}{l}{\textit{C. Transplant (Plan Transfer)}} \\
\hspace{3mm} Latent Sequence & 2.5 & -0.05 & \textbf{48.3} \\
\hspace{3mm} \color{gray}{Explicit CoT (Ref)} & \color{gray}{95.0} & \color{gray}{-4.51} & \color{gray}{30.0} \\
\bottomrule
\end{tabular}
\caption{\textbf{Causal Intervention Results.} Early ablation harms performance, while norm-preserving edits steer the outcome. Notably, latent transplants transfer "reasoning plans" (high Directional \%) better than explicit text.}
\label{tab:causal-interventions}
\end{table}

\paragraph{Results.} 
Table~\ref{tab:causal-interventions} summarizes the causal effects. First, ablation experiments (Figure~\ref{fig:ablation}) show that early latent vectors are critical; removing them causes a sharp drop in performance. Second, targeted edits prove that these vectors are steerable: small directional updates reliably shift the output probabilities and can flip the final answer. Finally, the transplant experiment demonstrates effective "plan transfer." Unlike explicit CoT, transplanting latent sequences successfully biases the model toward the donor's answer. Together, these results confirm that latent sequences encode robust and transferable reasoning trajectories.

\subsection{Uncovering Architectural and Geometric Underpinnings}
\label{sec:arch-geom}

What properties of the Transformer architecture enable this stable, structured latent computation? We hypothesize two key factors: the geometry induced by tied input-output embeddings and the emergence of a monotonically decreasing energy function along the latent chain.

\paragraph{Weight Tying as a Nearest-Neighbor Projector.}
In models with tied embeddings, the output projection matrix is the transpose of the input embedding matrix ($W_o = E^\top$). Consequently, selecting the next token $\hat{\imath}$ at low temperature is equivalent to finding the token embedding $\mathbf{e}_{\hat{\imath}}$ with the highest cosine similarity to the final hidden state $\mathbf{h}$:
\begin{equation}
\hat{\imath} = \arg\max_i \langle \mathbf{h}, \mathbf{e}_i \rangle.
\end{equation}
The standard "decode-then-re-embed" cycle can thus be viewed as a nearest-neighbor projection $\mathcal{T}(\mathbf{h}) = \mathbf{e}_{\hat{\imath}}$, which maps $\mathbf{h}$ onto the discrete codebook of token embeddings. Our latent update mechanism, which directly feeds $\mathbf{h}$ as the next input, is a continuous approximation of this cycle. This approximation is accurate when $\mathbf{h}$ is already close to an embedding vector $\mathbf{e}_{\hat{\imath}}$, i.e., when $\|\mathbf{h} - \mathcal{T}(\mathbf{h})\|^2$ is small. Untied embeddings introduce a tying gap, weakening this geometric correspondence and potentially destabilizing latent computation.

\paragraph{A Monotone Energy Over the Latent Chain.}
We further hypothesize that the latent trajectory follows a directed path from a high-energy "unresolved" state to a low-energy "resolved" state. We test this by learning a simple scalar energy function $H:\mathbb{R}^d \to \mathbb{R}$ optimized with a margin ranking loss to ensure energy decreases at each step along the chain from the last question token to the first answer token:
\begin{equation}
\label{eq:rank}
\mathcal{L}_{\text{rank}} = \mathbb{E} \left[ \max\bigl\{0, \gamma - (H(\mathbf{x}_t)-H(\mathbf{x}_{t+1}))\bigr\} \right].
\end{equation}

\paragraph{Results.}
Table~\ref{tab:architectural-analysis} validates both architectural hypotheses. First, \textbf{weight tying} proves critical; models with tied embeddings achieve significantly higher accuracy ratios ($r_{\text{Acc}}$) compared to untied variants. Second, the learned energy landscape exhibits monotonicity along the latent chain, with rank correlations approaching 1.0. This confirms that the latent process is not a random walk, but a \textbf{structured, contractive descent} toward the final answer.

\begin{table}[t]
\centering
\small
\setlength{\tabcolsep}{5pt}
\begin{tabular}{lr}
\toprule
\textbf{Architectural Property} & \textbf{Result} \\
\midrule
\multicolumn{2}{l}{\textit{\textbf{I. Impact of Weight Tying}}} \\
Accuracy Ratio ($r_{\text{Acc}}$) -- \textbf{Tied (Ours)} & \textbf{2.97} \\
Accuracy Ratio ($r_{\text{Acc}}$) -- Untied & 1.15 \\
\addlinespace[5pt]
\multicolumn{2}{l}{\textit{\textbf{II. Energy Landscape Monotonicity}}} \\
Strictly Monotonic Chains & \textbf{99.0\%} \\
Spearman Rank Correlation & $>0.99$ \\
\bottomrule
\end{tabular}
\caption{\textbf{Geometric Underpinnings.} Weight tying is crucial for latent efficacy ($r_{\text{Acc}}$), and the learned energy function confirms a highly structured, monotonic descent towards the solution.}
\label{tab:architectural-analysis}
\end{table}

In summary, this section establishes a unified interpretability framework for the continuous-thought paradigm. We demonstrate that latent vectors are not arbitrary states, but semantically aligned representations that precise causal control over the thinking process. Our findings validate latent reasoning as a robust and interpretable alternative to explicit CoT.

\section{Interventions on Latent Reasoning}
\label{sec:intervene}

Building upon the interpretability analysis in Section~\ref{sec:latent-interpretation}, we now operationalize our findings. We introduce a series of novel training-free interventions that manipulate cached latent vectors at decode-time. Formally, given a latent state $\mathbf{z}_k$, we seek a steered state $\mathbf{z}'_k$ that maximizes alignment with a prior-defined reasoning manifold while minimizing deviation from the original context. As detailed in Algorithm~\ref{alg:unified-steering}, this process is governed by a generalized update rule, $\mathbf{z}'_k \leftarrow \mathcal{I}(\mathbf{z}_k; \Phi, \Omega)$, where the \textit{guidance prior} $\Phi$ and \textit{manifold constraint} $\Omega$ are instantiated based on the specific interpretability source (Semantic, Causal, or Geometric). Our goal is to demonstrate that by leveraging a theoretical understanding of the latent reasoning process, we can unlock and enhance the model's inherent inferential capabilities without any parameter updates.

\subsection{Experimental Setup}
\label{sec:setup}

\paragraph{Latent Reasoning Paradigms.}
We evaluate our interventions on two distinct paradigms within the continuous-thought framework: CoConut employs a latent span optimization to fill reserved placeholders with continuous thoughts, CODI distills a long, explicit Chain-of-Thought into a compressed sequence of continuous vectors prefixed to the input.

\paragraph{Models and Tasks.}
We conduct experiments across Qwen3-8B, Llama-3.1-8B and the smaller Llama-3.2-3B. To rigorously evaluate efficacy and scalability, Our evaluation covers three distinct domains:
(1) \textbf{In-Domain Mathematical Reasoning}: We use the standard GSM8K~\citep{gsm8k} test set.
(2) \textbf{Out-of-Domain (OOD) Robustness}: We evaluate on GSM-Hard~\citep{gsmhard} and SVAMP~\citep{svamp} to test generalization to higher complexity and varying linguistic patterns.
(3) \textbf{Implicit Commonsense Reasoning}: We include StrategyQA~\citep{strategyqa} to verify that our findings extend beyond mathematical tasks.

\subsection{Intervention A: Semantic Structure Transport}
\label{subsec:intervene-semantic}

Drawing from our discovery in Section~\ref{sec:probe} that latent vectors are semantically aligned with explicit CoT steps, we propose \textbf{Mapper-Guided Transport} to refine the latent trajectory. We identify the mean $\mathbf{m}_0$ of the terminal latent pair and map it to a target semantic destination $\mathbf{t}_0 = F(\mathbf{m}_0)$ using the linear mapper trained in Section 3.

We then steer the latent state using spherical linear interpolation (slerp) and norm mixing:
\begin{equation}
\label{eq:mos}
\begin{aligned}
\widehat{\mathbf{u}} &= \operatorname{slerp}\!\Big( \frac{\mathbf{m}_0}{\|\mathbf{m}_0\|}, \frac{\mathbf{t}_0}{\|\mathbf{t}_0\|}; \alpha \Big), \\
\widehat{\mathbf{m}} &= \Big[(1-\eta)\|\mathbf{m}_0\|+\eta\|\mathbf{t}_0\|\Big]\;\widehat{\mathbf{u}},
\end{aligned}
\end{equation}
where $\alpha$ controls the directional shift and $\eta$ regulates the norm influence. To preserve instance-specific information, we re-apply the original residual vectors to the updated mean. Finally, to ensure the modified vectors remain on the valid token manifold, we apply \textbf{Embedding-Subspace Alignment} via a projector $P_r$ derived from the principal components of the embedding matrix.

\begin{algorithm}[t]
\caption{Unified Interpretability-Guided Latent Steering}
\label{alg:unified-steering}
\begin{algorithmic}[1]

\Require Latent State $\mathbf{z}_k$, Variant $\mathcal{M}$, Hyperparams $\alpha, \lambda, \eta$

\Statex \hspace{-2em} \textbf{Configuration} (Define Prior $\Phi$ and Constraint $\Omega$ via $\mathcal{M}$)

\Statex $\begin{aligned}
\textbf{\small A:} \quad & \Phi(\mathbf{z}) = F_{\text{map}}(\mathbf{z}), & \Omega(\mathbf{z}) &= P_{\text{subspace}}\mathbf{z} \\
\textbf{\small B:} \quad & \Phi(\mathbf{z}) = \mathbf{z} - \eta\nabla\mathcal{L}, & \Omega(\mathbf{z}) &= \mathbf{z} / \|\mathbf{z}\| \\
\textbf{\small C:} \quad & \Phi(\mathbf{z}) = \mathbf{z} - \eta\nabla H, & \Omega(\mathbf{z}) &= \text{WT-Proj}(\mathbf{z})
\end{aligned}$

\Statex \hspace{-2em} \textbf{Execution}

\State $\mathbf{v}^* \gets \Phi(\mathbf{z}_k)$ \Comment{Extract guidance signal}
\State $\mathbf{d} \gets \mathbf{v}^* - \mathbf{z}_k$ 
\State $\mathbf{z}^{\text{steer}} \gets \mathbf{z}_k + \alpha \cdot \frac{\mathbf{d}}{\|\mathbf{d}\|} \cdot \|\mathbf{z}_k\|$ \Comment{Apply directional update}
\State $\mathbf{z}'_k \gets (1-\lambda)\mathbf{z}^{\text{steer}} + \lambda \cdot \Omega(\mathbf{z}^{\text{steer}})$ \Comment{Manifold regularization}

\State \Return $\mathbf{z}'_k$
\end{algorithmic}
\end{algorithm}

\begin{table*}[t]
\centering
\scriptsize
\setlength{\tabcolsep}{3.5pt} 
\renewcommand{\arraystretch}{1.15} 
\begin{tabular}{@{}ll ccc c ccc c ccc@{}}
\toprule
& & \multicolumn{3}{c}{\textbf{Qwen3-8B}} & & \multicolumn{3}{c}{\textbf{Llama-3.1-8B}} & & \multicolumn{3}{c}{\textbf{Llama-3.2-3B}} \\
\cmidrule(lr){3-5} \cmidrule(lr){7-9} \cmidrule(lr){11-13}
& & \multicolumn{1}{c}{\textbf{In-domain}} & \multicolumn{2}{c}{\textbf{OOD}} & & \multicolumn{1}{c}{\textbf{In-domain}} & \multicolumn{2}{c}{\textbf{OOD}} & & \multicolumn{1}{c}{\textbf{In-domain}} & \multicolumn{2}{c}{\textbf{OOD}} \\
\cmidrule(lr){3-3} \cmidrule(lr){4-5} \cmidrule(lr){7-7} \cmidrule(lr){8-9} \cmidrule(lr){11-11} \cmidrule(lr){12-13}
\textbf{Variant} & \textbf{Paradigm} & \textbf{GSM8K} & \textbf{GSM-H} & \textbf{SVAMP} & & \textbf{GSM8K} & \textbf{GSM-H} & \textbf{SVAMP} & & \textbf{GSM8K} & \textbf{GSM-H} & \textbf{SVAMP} \\
\midrule

\multicolumn{13}{l}{\textit{Baselines}} \\
\multirow{2}{*}{No Interv.} & CoConut & 50.4 & 14.1 & 68.5 & & 44.8 & 12.2 & 63.2 & & 42.1 & 11.5 & 58.9 \\
& CODI & 62.1 & 17.9 & 80.3 & & 60.7 & 14.5 & 76.4 & & 58.9 & 13.4 & 72.3 \\
\midrule

\multicolumn{13}{l}{\textbf{A. Semantic Structure} (Sec.~\ref{subsec:intervene-semantic})} \\
\multirow{2}{*}{Mapper} & CoConut & 51.9 \tiny{(+1.5)} & 15.3 \tiny{(+1.2)} & 69.6 \tiny{(+1.1)} & & 46.2 \tiny{(+1.4)} & 13.5 \tiny{(+1.3)} & 64.2 \tiny{(+1.0)} & & 43.2 \tiny{(+1.1)} & 12.4 \tiny{(+0.9)} & 60.1 \tiny{(+1.2)} \\
& CODI & \textbf{63.5} \tiny{(+1.4)} & 18.9 \tiny{(+1.0)} & \textbf{81.6} \tiny{(+1.3)} & & \textbf{62.2} \tiny{(+1.5)} & \textbf{15.9} \tiny{(+1.4)} & 77.7 \tiny{(+1.3)} & & 59.6 \tiny{(+0.7)} & \textbf{14.7} \tiny{(+1.3)} & \textbf{73.6} \tiny{(+1.3)} \\
\midrule

\multicolumn{13}{l}{\textbf{B. Causal Hub Editing} (Sec.~\ref{subsec:intervene-causal})} \\
\multirow{2}{*}{\textit{Slerp (B.1)}} & CoConut & 51.6 \tiny{(+1.2)} & \textbf{15.5} \tiny{(+1.4)} & \textbf{69.7} \tiny{(+1.2)} & & 46.2 \tiny{(+1.4)} & 13.3 \tiny{(+1.1)} & \textbf{64.4} \tiny{(+1.2)} & & 43.0 \tiny{(+0.9)} & 12.6 \tiny{(+1.1)} & 59.6 \tiny{(+0.7)} \\
& CODI & 62.5 \tiny{(+0.4)} & 18.6 \tiny{(+0.7)} & 81.4 \tiny{(+1.1)} & & 61.3 \tiny{(+0.6)} & 15.4 \tiny{(+0.9)} & \textbf{77.8} \tiny{(+1.4)} & & 59.8 \tiny{(+0.9)} & 14.4 \tiny{(+1.0)} & 73.3 \tiny{(+1.0)} \\

\multirow{2}{*}{\textit{Gradient (B.2)}} & CoConut & \textbf{52.2} \tiny{(+1.8)} & 15.1 \tiny{(+1.0)} & 69.4 \tiny{(+0.9)} & & \textbf{46.4} \tiny{(+1.6)} & 13.1 \tiny{(+0.9)} & 64.2 \tiny{(+1.0)} & & 43.3 \tiny{(+1.2)} & \textbf{12.7} \tiny{(+1.2)} & 60.0 \tiny{(+1.1)} \\
& CODI & 62.9 \tiny{(+0.8)} & 19.2 \tiny{(+1.3)} & 81.1 \tiny{(+0.8)} & & 61.1 \tiny{(+0.4)} & 15.7 \tiny{(+1.2)} & 77.0 \tiny{(+0.6)} & & 59.7 \tiny{(+0.8)} & 14.2 \tiny{(+0.8)} & 73.4 \tiny{(+1.1)} \\
\midrule

\multicolumn{13}{l}{\textbf{C. Geometric Priors} (Sec.~\ref{subsec:intervene-arch})} \\
\multirow{2}{*}{\textit{WT-Proj (C.1)}} & CoConut & 51.0 \tiny{(+0.6)} & 14.9 \tiny{(+0.8)} & 69.5 \tiny{(+1.0)} & & 45.5 \tiny{(+0.7)} & 13.0 \tiny{(+0.8)} & 64.1 \tiny{(+0.9)} & & 43.1 \tiny{(+1.0)} & 12.3 \tiny{(+0.8)} & \textbf{60.1} \tiny{(+1.2)} \\
& CODI & 62.8 \tiny{(+0.7)} & 18.8 \tiny{(+0.9)} & 81.3 \tiny{(+1.0)} & & 61.6 \tiny{(+0.9)} & 15.3 \tiny{(+0.8)} & 77.6 \tiny{(+1.2)} & & 59.6 \tiny{(+0.7)} & 14.4 \tiny{(+1.0)} & 72.7 \tiny{(+0.4)} \\

\multirow{2}{*}{\textit{Energy (C.2)}} & CoConut & 51.4 \tiny{(+1.0)} & 15.2 \tiny{(+1.1)} & 69.4 \tiny{(+0.9)} & & 45.4 \tiny{(+0.6)} & 13.1 \tiny{(+0.9)} & \textbf{64.4} \tiny{(+1.2)} & & 42.9 \tiny{(+0.8)} & \textbf{12.7} \tiny{(+1.2)} & 59.8 \tiny{(+0.9)} \\
& CODI & 62.6 \tiny{(+0.5)} & \textbf{19.3} \tiny{(+1.4)} & 81.5 \tiny{(+1.2)} & & 61.9 \tiny{(+1.2)} & 15.7 \tiny{(+1.2)} & 77.5 \tiny{(+1.1)} & & 59.8 \tiny{(+0.9)} & 14.5 \tiny{(+1.1)} & 73.3 \tiny{(+1.0)} \\

\bottomrule
\end{tabular}
\caption{
\textbf{Unified Performance Analysis on Mathematical Reasoning.} We report the accuracy (\%) on GSM8K (In-Domain), GSM-Hard and SVAMP (OOD) across three model scales. The results are grouped by intervention family (A, B, C). Boldface highlights the best performance for each model-paradigm configuration.
}
\label{tab:unified-math}
\end{table*}

\subsection{Intervention B: Causal Hub Editing}
\label{subsec:intervene-causal}

Our causal analysis in Section~\ref{sec:causal-latents} identified early latent vectors (e.g., $\mathbf{z}_2$) as "causal hubs" that exert disproportionate control over the final output. We introduce two strategies to strictly edit these critical states.

\paragraph{Answer-Anchored Slerp.}
We reorient the critical vector $\mathbf{z}_2$ toward a boundary vector $\mathbf{t}$ (derived from a retrieved exemplar) to steer the reasoning direction:
\begin{equation}
\label{eq:s1}
\mathbf{z}_2' = \|\mathbf{z}_2\| \; \operatorname{slerp}\!\Big( \frac{\mathbf{z}_2}{\|\mathbf{z}_2\|}, \frac{\mathbf{t}}{\|\mathbf{t}\|}; \alpha \Big).
\end{equation}

\paragraph{Answer-Directed Gradient Update.}
Alternatively, we apply a single gradient step to maximize the likelihood of the correct answer direction while strictly enforcing a norm constraint to prevent energy explosion:
\begin{equation}
\label{eq:s2}
\mathbf{z}_2' = \mathbf{z}_2 - \eta\,\|\mathbf{z}_2\| \; \frac{\nabla_{\mathbf{z}_2}\mathcal{L}}{\|\nabla_{\mathbf{z}_2}\mathcal{L}\|+\varepsilon}.
\end{equation}

\subsection{Intervention C: Geometric Priors}
\label{subsec:intervene-arch}

Based on the architectural findings in Section~\ref{sec:arch-geom}, we propose interventions that enforce the geometric consistency of the latent space.

\paragraph{Weight-Tying Consistent Projection.}
To strengthen the link between hidden states and the output vocabulary space, we nudge the hidden state $\mathbf{h}_\ell$ towards its expected embedding representation $\tilde{\mathbf{e}}_\ell$:
\begin{equation}
\label{eq:wt}
\tilde{\mathbf{e}}_\ell = E^{\top}\mathrm{softmax}(\mathbf{o}_{\ell-1}/\tau), \quad \mathbf{h}_\ell' = (1-\alpha)\mathbf{h}_\ell + \alpha\,\tilde{\mathbf{e}}_\ell.
\end{equation}

\paragraph{Energy-Guided Local Descent.}
Using the monotonicity energy function $H(\cdot)$ trained in Section 3, we apply a trust-region gradient descent step. This ensures the reasoning process progresses "downhill" in the energy landscape, stabilizing the inference trajectory:
\begin{equation}
\label{eq:energy}
\mathbf{h}_\ell' = \mathbf{h}_\ell - \operatorname{Proj}_{\mathbb{B}(\mathbf{0},\rho\|\mathbf{h}_\ell\|)} \!\Big(\eta\,\nabla H(\mathbf{h}_\ell)\Big).
\end{equation}

\subsection{Experimental Results and Analysis}
\label{sec:results}

We present a unified evaluation of our interventions. Table~\ref{tab:unified-math} summarizes the performance across all mathematical reasoning benchmarks (In-Domain and OOD), organized hierarchically by intervention family. Table~\ref{tab:strategyqa} details the results on commonsense reasoning.

\paragraph{Efficacy on Mathematical Reasoning.}
Table~\ref{tab:unified-math} demonstrates that our interpretability-guided interventions yield consistent improvements across all settings. Semantic Transport (Method A) proves particularly robust for OOD tasks, validating that aligning latents with semantic CoT clusters improves generalization. Causal Hub Editing (Method B) is highly effective as shown, the Gradient-based update (B.2) achieves the highest gains on the standard GSM8K benchmark, corroborating our finding that early latent vectors act as steerable control points. Geometric Priors (Method C) offer stable gains, confirming that enforcing architectural constraints regularizes the inference process effectively.

\paragraph{Scalability to Model-Scales.}
Table~\ref{tab:unified-math} show that our interventions remain highly effective cross different scales. For instance, Mapper-Guided Transport improves CODI performance on GSM-Hard by +1.3\%. This indicates that the latent structures we exploit are fundamental to the paradigm.

\paragraph{Generalization to Commonsense Reasoning.}
We report results on StrategyQA in Table~\ref{tab:strategyqa}in order to assess domain universality. All interventions provide positive gains, with Causal Hub Editing showing particular strength (up to +1.5\% for Qwen-CODI). This suggests that the causal mechanisms are task-agnostic properties of continuous thought.

\begin{table}[t]
\centering
\scriptsize
\setlength{\tabcolsep}{3pt} 
\renewcommand{\arraystretch}{1.15}
\begin{tabular}{@{}ll c c c c c@{}}
\toprule
& & \textbf{Qwen3-8B} & & \textbf{Llama-3.1-8B} & & \textbf{Llama-3.2-3B} \\
\cmidrule(lr){3-3} \cmidrule(lr){5-5} \cmidrule(lr){7-7}
\textbf{Variant} & \textbf{Paradigm} & \textbf{StrategyQA} & & \textbf{StrategyQA} & & \textbf{StrategyQA} \\
\midrule

\multicolumn{7}{l}{\textit{Baselines}} \\
\multirow{2}{*}{No Interv.} & CoConut & 75.5 & & 71.4 & & 67.6 \\
& CODI & 80.9 & & 76.3 & & 70.2 \\
\midrule

\multicolumn{7}{l}{\textbf{A. Semantic Structure}} \\
\multirow{2}{*}{Mapper} & CoConut & 76.6 \tiny{(+1.1)} & & 72.4 \tiny{(+1.0)} & & 68.3 \tiny{(+0.7)} \\
& CODI & 82.1 \tiny{(+1.2)} & & 77.4 \tiny{(+1.1)} & & \textbf{71.3} \tiny{(+1.1)} \\
\midrule

\multicolumn{7}{l}{\textbf{B. Causal Hub Editing}} \\
\multirow{2}{*}{\textit{Slerp (B.1)}} & CoConut & 76.7 \tiny{(+1.2)} & & 72.6 \tiny{(+1.2)} & & \textbf{68.8} \tiny{(+1.2)} \\
& CODI & \textbf{82.4} \tiny{(+1.5)} & & 77.1 \tiny{(+0.8)} & & 71.1 \tiny{(+0.9)} \\
\addlinespace[3pt]
\multirow{2}{*}{\textit{Gradient (B.2)}} & CoConut & \textbf{76.8} \tiny{(+1.3)} & & \textbf{72.7} \tiny{(+1.3)} & & 68.5 \tiny{(+0.9)} \\
& CODI & 81.9 \tiny{(+1.0)} & & \textbf{77.5} \tiny{(+1.2)} & & 71.2 \tiny{(+1.0)} \\
\midrule

\multicolumn{7}{l}{\textbf{C. Geometric Priors}} \\
\multirow{2}{*}{\textit{WT-Proj (C.1)}} & CoConut & 76.4 \tiny{(+0.9)} & & 72.2 \tiny{(+0.8)} & & 68.7 \tiny{(+1.1)} \\
& CODI & 81.9 \tiny{(+1.0)} & & 77.3 \tiny{(+1.0)} & & \textbf{71.3} \tiny{(+1.1)} \\
\addlinespace[3pt]
\multirow{2}{*}{\textit{Energy (C.2)}} & CoConut & 76.1 \tiny{(+0.6)} & & 72.4 \tiny{(+1.0)} & & 68.0 \tiny{(+0.4)} \\
& CODI & 82.2 \tiny{(+1.3)} & & 77.2 \tiny{(+0.9)} & & \textbf{71.3} \tiny{(+1.1)} \\

\bottomrule
\end{tabular}
\caption{
\textbf{Impact on Commonsense Reasoning (StrategyQA).} We apply the same interventions to evaluate cross-domain applicability. Values report accuracy (\%) with absolute improvement in parentheses.
}
\label{tab:strategyqa}
\end{table}

\section{Conclusion}
This paper establishes a unified framework of interpretability and intervention in latent reasoning. We demonstrate that continuous thought vectors are not artifacts but structured semantic trajectories governed by distinct geometric and causal priors. Translating these insights into training-free interventions leads to consistent performance gains across mathematics, commonsense domains, and different model scales. These improvements serve as a strong validation of our interpretability hypotheses, demonstrating that latent reasoning is interpretably controllable, offering a robust foundation for future latent reasoning paradigms.

\section*{Limitations}
While our training-free interventions significantly enhance reasoning performance without parameter updates, they introduce a marginal computational overhead during inference due to the additional calculation of gradients and projections. Future work aims to eliminate this latency by incorporating these interpretability priors as supervisory signals, thereby developing improved training algorithms for latent reasoning.

\section*{Acknowledgments}
The work was supported by the National Natural Science Foundation of China (Grant No. 62471287).


\bibliography{custom}

\newpage
\appendix

\section{Detailed Experimental Setup}
\label{sec:app-setup}

\begin{table*}[h]
\centering

\setlength{\tabcolsep}{10pt} 
\renewcommand{\arraystretch}{1.25} 

\begin{tabular}{@{}l l c c c@{}}
\toprule
\textbf{Intervention Method} & \textbf{Hyperparameter} & \textbf{Search Range} & \textbf{Optimal} & \textbf{Acc. (\%)} \\
\midrule

\textit{Baseline (No Interv.)} & \multicolumn{1}{c}{--} & \multicolumn{1}{c}{--} & \multicolumn{1}{c}{--} & 50.4 \\
\midrule

\multicolumn{5}{l}{\textbf{A. Semantic Structure}} \\
\multirow{2}{*}{Mapper-Guided} 
    & Slerp factor $\alpha$ & $[0.05, 0.25]$ & $0.15$ & \multirow{2}{*}{51.9} \\
    & Norm mix $\eta$       & fixed & $0.25$ & \\
\midrule

\multicolumn{5}{l}{\textbf{B. Causal Hub Editing}} \\

Anchored Slerp (B.1) 
    & Slerp factor $\alpha$ & $[0.05, 0.25]$ & $0.10$ & 51.6 \\
\cmidrule(l{0.5em}){2-5}
Gradient Edit (B.2)
    & Step size $\eta$      & $[0.05, 0.25]$ & $0.20$ & \textbf{52.2} \\

\midrule

\multicolumn{5}{l}{\textbf{C. Geometric Priors}} \\
\multirow{2}{*}{WT-Proj (C.1)} 
    & Mix ratio $\alpha$    & $[0.05, 0.25]$ & $0.20$ & \multirow{2}{*}{51.0} \\
    & Temp. $\tau$          & fixed          & $1.0$  & \\
\cmidrule(l{0.5em}){2-5}
\multirow{2}{*}{Energy Descent (C.2)} 
    & Step size $\eta$      & $[1\text{e-}3, 5\text{e-}3]$ & $2\text{e-}3$ & \multirow{2}{*}{51.4} \\
    & Trust region $\rho$   & fixed & $0.25$ & \\

\bottomrule
\end{tabular}
\caption{\textbf{Hyperparameter sensitivity analysis.} Grid search results on the Qwen3-8B (CoConut) model using the GSM8K validation set. We identify optimal configurations for each intervention family.}
\label{tab:hyperparam-search}
\end{table*}

\subsection{Model Configurations and Training}
\label{subsec:app-config}
We conduct our experiments using the CoConut \citep{coconut} and CODI \citep{codi} paradigms. To ensure a fair comparison and reproducibility, we strictly adhere to the model architectures and training protocols specified in their respective official repositories\footnote{We refer to the official implementations of CoConut and CODI.}, with specific adjustments for model scaling.

\begin{figure*}[h]
    \centering
    \includegraphics[width=\textwidth]{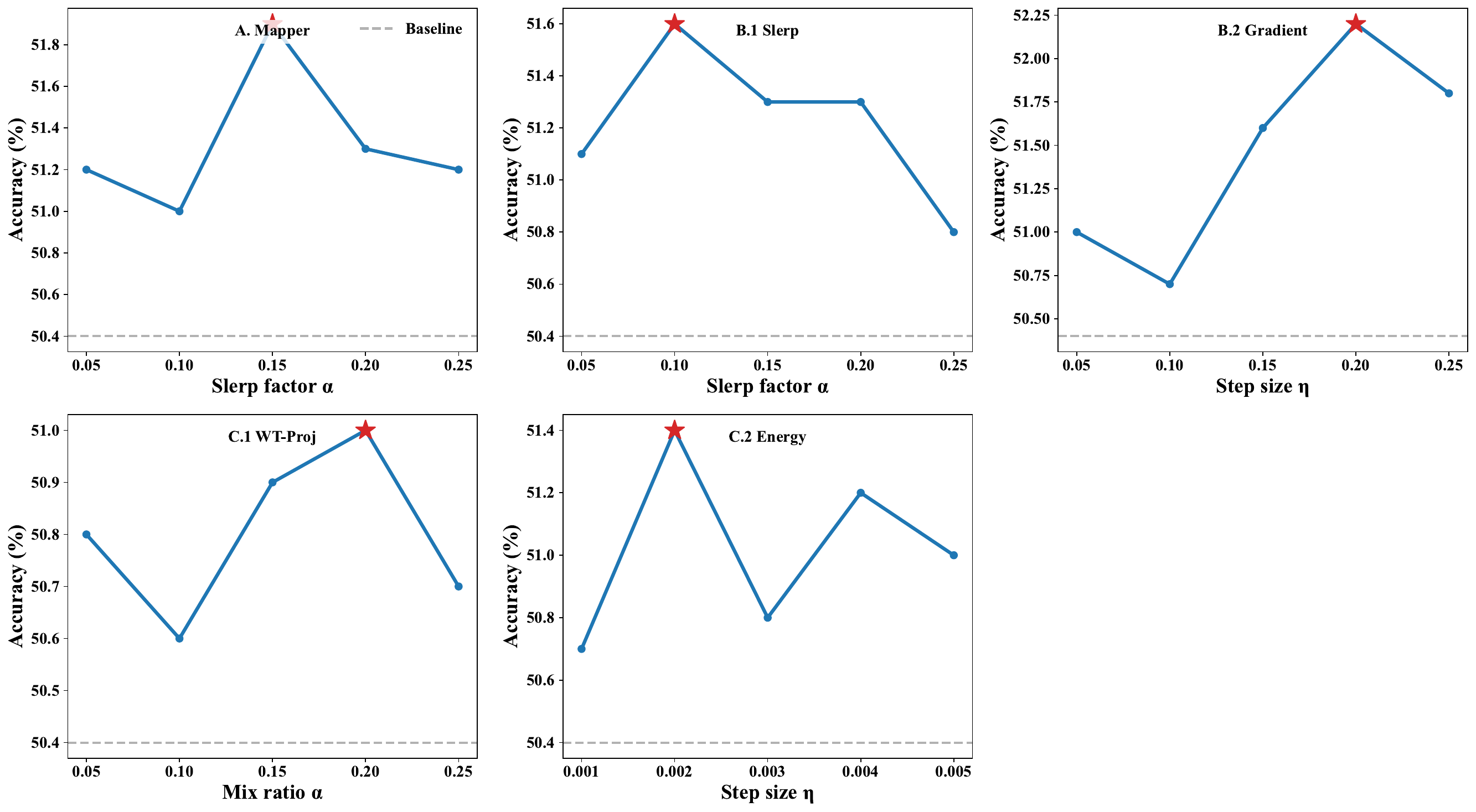}
    \caption{\textbf{Hyperparameter sensitivity analysis.} Each subplot varies one specific hyperparameter. The red stars indicate the optimal configurations. The baselines represent performance without intervention.}
    \label{fig:hyperparam-plots}
\end{figure*}

\paragraph{Hyperparameters.}

For all experiments, we maintain the latent span length at $K=6$. We use the AdamW optimizer with $\beta_1=0.9$ and $\beta_2=0.95$. The specific learning rate schedules differ by model size to ensure stability: For 8B Models (Qwen3-8B, Llama-3.1-8B), We use a peak learning rate of $1\times 10^{-5}$ with cosine annealing schedule. For Llama-3.2-3B, we use a peak learning rate of $2\times 10^{-5}$. The training batch size is set to 16.

All other hyperparameters, including warm-up steps, epochs, and distillation loss weights for CODI, are kept the same as the default settings in the original papers.

\subsection{Compute Resources}
All training and inference experiments were conducted on a cluster equipped with 8 NVIDIA A100 (80GB) GPUs. Under this hardware configuration, training the latent reasoning models (both CoConut and CODI variants) for the 3B and 8B models required approximately 24 hours.

\section{Details of Interpretability Probes}
\label{sec:app-probes}

\subsection{Probe Architectures and Training}
To analyze the semantic content of latent vectors (Section~\ref{sec:probe}), we trained specific probe networks.

\paragraph{Dataset Construction.}
We constructed a probing dataset using the GSM8K training set. We randomly sampled 1,000 instances and generated both the latent thought sequences and the ground-truth explicit Chain-of-Thought (CoT) paths. This resulted in a paired dataset of aligned latent vectors and text representations.

\paragraph{Linear Mapper ($f_{\phi}$).}
The mapper is designed to test linear recoverability. It is a simple transformation $f_{\phi}: \mathbb{R}^{d} \to \mathbb{R}^{d}$, where $d$ is the model's hidden dimension. It is trained to minimize the cosine distance between the mapped latent vector and the corresponding CoT hidden state.

\paragraph{Energy Function ($H$).}
The monotonicity energy function $H(\cdot)$ is parameterized as a two-layer Multi-Layer Perceptron (MLP). The architecture consists of:
\begin{equation}
    \text{Input} \xrightarrow{W_1} \text{Hidden} \xrightarrow{\text{ReLU}} \text{Hidden} \xrightarrow{W_2} \text{Output}
\end{equation}
The hidden dimension is set to equal the model's embedding dimension. We train this network using the margin ranking loss described in Eq.~\ref{eq:rank} to assign lower energy values to latent states closer to the solution.

\subsection{Details on Structural Probes (CKA)}
For the Centered Kernel Alignment (CKA) analysis, we employed Linear CKA to measure the similarity between the geometry of the latent space and the explicit CoT space. We computed the similarity matrix using a batch size of 64 reasoning steps sampled from the probing dataset, ensuring a robust estimation of the structural correspondence.

\section{Intervention Implementation Details}
\label{sec:app-intervention}

\subsection{Hyperparameter Sensitivity and Selection}
Our interventions involve specific hyperparameters that control the strength of the edit (e.g., interpolation factor $\alpha$, step size $\eta$). We performed a grid search to identify optimal values on a held-out validation set.

Table~\ref{tab:hyperparam-search} presents the sensitivity analysis for the Qwen3-8B model trained with CoConut on the GSM8K In-domain task. The "Optimal" column corresponds to the results reported in the main paper. Crucially, the optimal hyperparameters identified here were kept fixed and applied consistently across all other model scales (Llama-3.1-8B, Llama-3.2-3B) and datasets (OOD, StrategyQA) to demonstrate the robustness and transferability of our approach.

\begin{figure*}[h]
    \centering

    \begin{subfigure}{\textwidth}
        \centering
        \includegraphics[width=\textwidth]{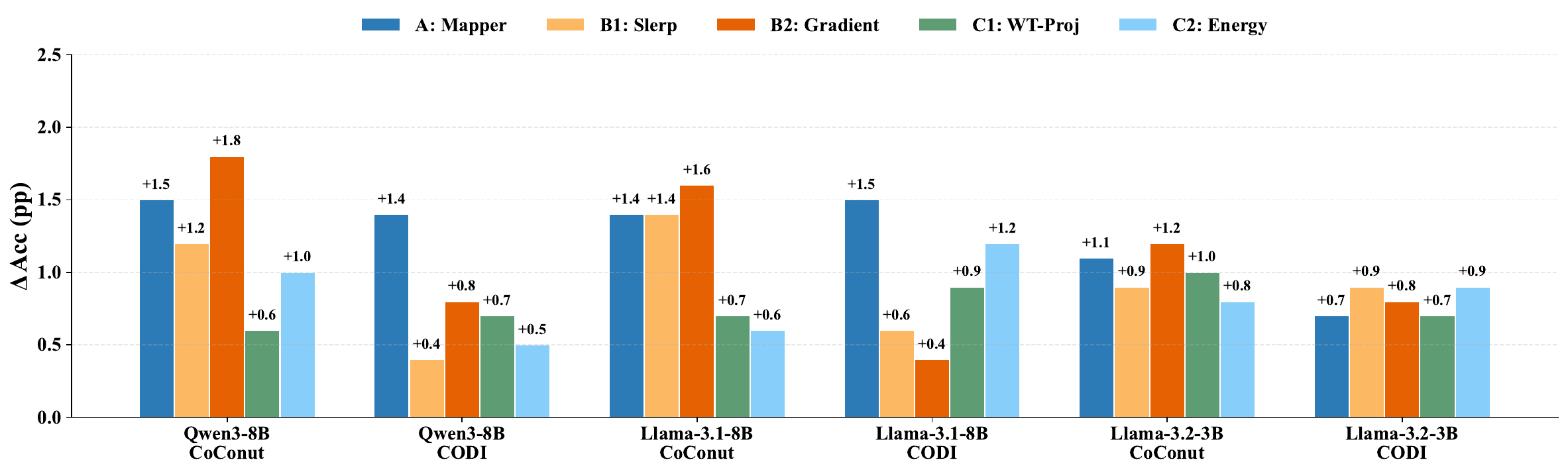}
        \caption{In-Domain: GSM8K}
        \label{fig:math-gsm8k}
    \end{subfigure}

    \begin{subfigure}{\textwidth}
        \centering
        \includegraphics[width=\textwidth]{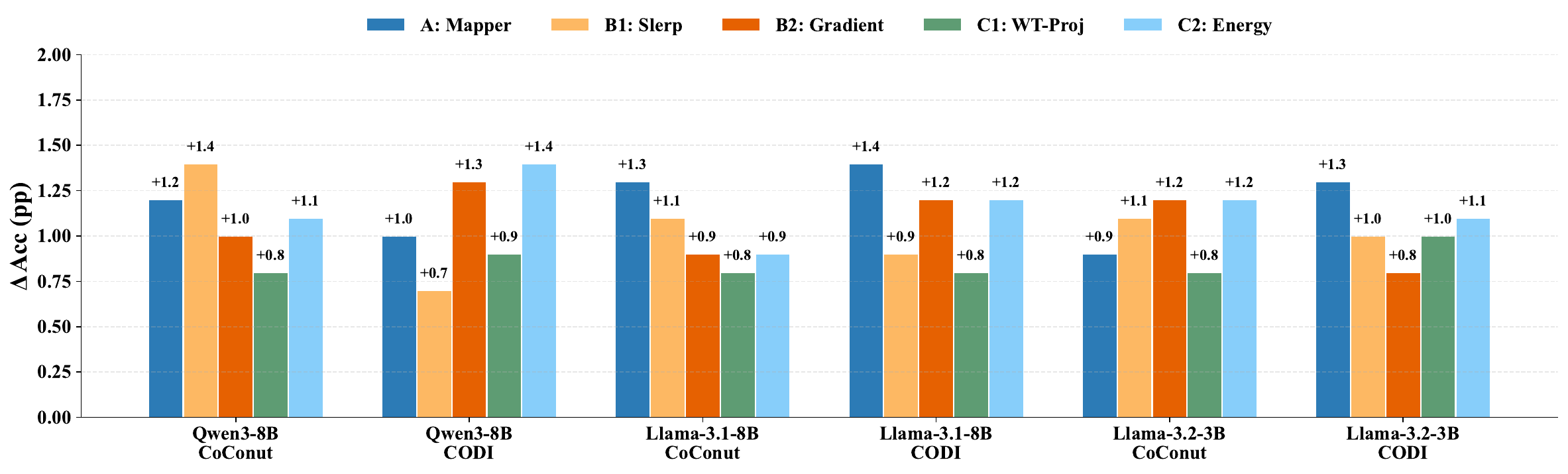}
        \caption{Out-of-Domain: GSM-Hard}
        \label{fig:math-gsmhard}
    \end{subfigure}

    \begin{subfigure}{\textwidth}
        \centering
        \includegraphics[width=\textwidth]{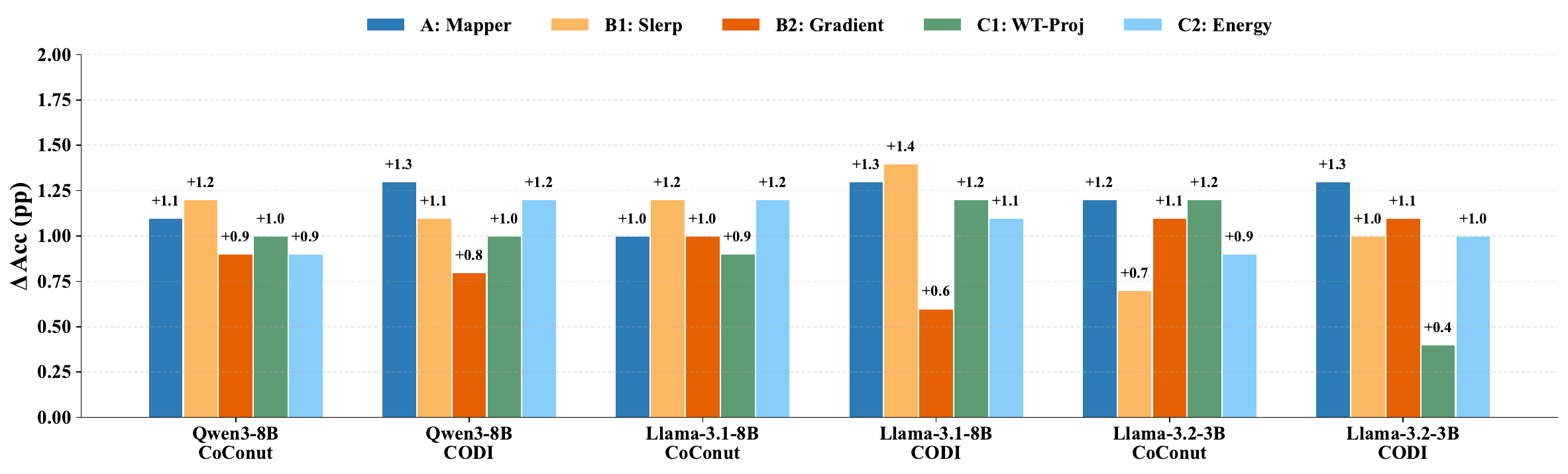}
        \caption{Out-of-Domain: SVAMP}
        \label{fig:math-svamp}
    \end{subfigure}
    
    \caption{\textbf{Unified Accuracy Improvements on Mathematical Reasoning.} Detailed breakdown of performance gains on (a) GSM8K, (b) GSM-Hard, and (c) SVAMP. By visualizing each dataset independently, we observe that Method B.2 (Gradient) excels in-domain, while Method A (Mapper) shows consistent robustness across OOD tasks.}
    \label{fig:math-results}
\end{figure*}

\begin{figure*}[h]
    \centering
    \includegraphics[width=\textwidth]{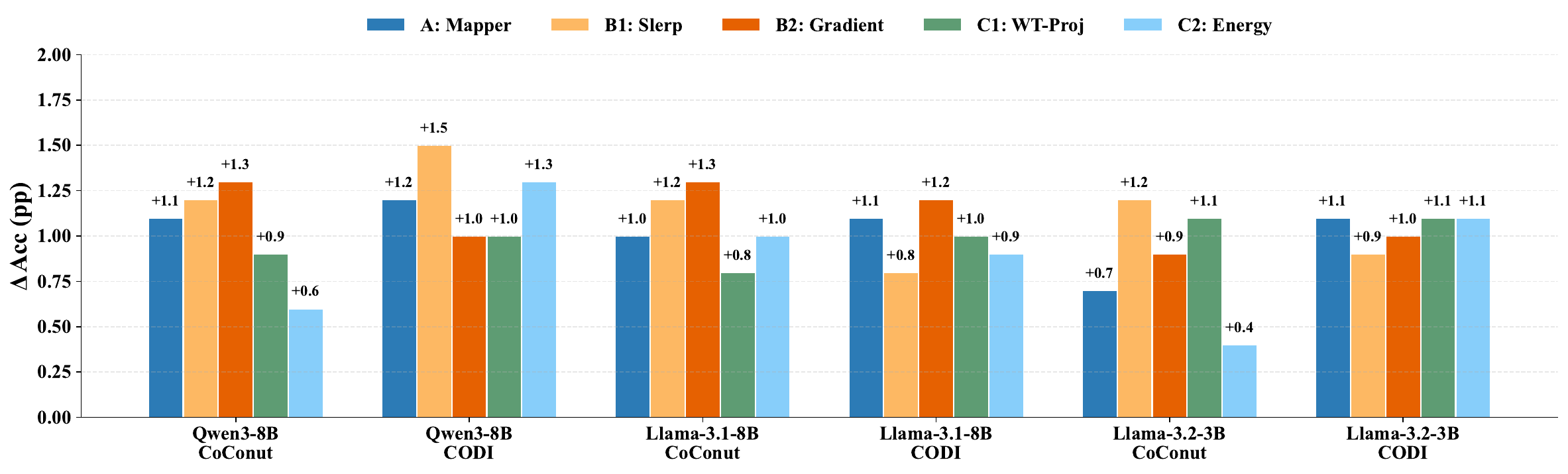}
    \caption{\textbf{Accuracy Improvements on StrategyQA (Commonsense Reasoning).} This visualization confirms that our interventions generalize effectively beyond mathematics, with consistent gains across model scales.}
    \label{fig:strategyqa-results}
\end{figure*}

\subsection{Retrieval Mechanism for Causal Editing}
For the Answer-Anchored Slerp intervention (Eq.~\ref{eq:s1}), we require a target vector $\mathbf{t}$ that represents a correct reasoning direction. We utilize the Latent LLM (CoConut or CODI) itself to encode the inputs. We extract the final hidden state of the input prompt to serve as the query vector, then compute the cosine similarity with the pre-cached latent vectors of the training exemplars. The top-1 retrieved instance ($k=1$) is selected, and its final latent thought vector is used as the anchor $\mathbf{t}$ to verify solution path within latent space.

\section{Additional Experimental Results}
\label{sec:app-results}

In this section, we provide supplementary visualizations to further substantiate our findings.

\subsection{Hyperparameter Sensitivity}
Figure~\ref{fig:hyperparam-plots} illustrates the impact of varying hyperparameter values on the in-domain GSM8K performance.

\subsection{Unified Performance Analysis}
We visualize the accuracy gains ($\Delta$ Accuracy) across all tasks. Figure~\ref{fig:math-results} details the improvements on mathematical reasoning benchmarks, separated by domain. Figure~\ref{fig:strategyqa-results} presents the results for commonsense reasoning.

\end{document}